\relax
\documentclass[letterpaper]{article} %DO NOT CHANGE THIS
\usepackage{arxiv}  %Required
\usepackage{times}  %Required
\usepackage{helvet}  %Required
\usepackage{courier}  %Required
\usepackage{url}  %Required
\usepackage{graphicx}  %Required
\usepackage{balance}
\usepackage[dvipsnames]{xcolor}

\usepackage{amsfonts}

\usepackage{tcolorbox}

\begin{document}
	% \title{Imagining an Engineer:\\
	% %The Perils of Data Bias Perpetuation Through GAN-based Data Augmentation}
	% %GAN-Based Data Augmentation Can Perpetuate Data (and Social) Biases}
	% On GAN-Based Data Augmentation Perpetuating Biases}
	% \author{Submission ID 119\\
	% \textbf{Disciplines:} AI, Computer Science, Society\\
	% %\textbf{Author Keywords:} Computer Vision, Generative Adversarial Networks, Social Bias, Gender Equality\\
	% %\textbf{Topics:} AI, risks, society
	% %Association for the Advancement of Artificial Intelligence\\
	% %2275 East Bayshore Road, Suite 160\\
	% %Palo Alto, California 94303\\
	% }
	
	\title{Imagining an Engineer:\\
		%The Perils of Data Bias Perpetuation Through GAN-based Data Augmentation}
		%GAN-Based Data Augmentation Can Perpetuate Data (and Social) Biases}
		On GAN-Based Data Augmentation Perpetuating Biases}
	\author{Niharika Jain$^*$ \and Lydia Manikonda$^*$ \and Alberto Olmo Hernandez$^*$ \AND Sailik Sengupta$^{* \dagger}$ \and Subbarao Kambhampati
		\\[0.5em]
		Arizona State University\\
		\{njain30, lmanikon, aolmoher, sailiks, rao\}@asu.edu
		%\textbf{Author Keywords:} Computer Vision, Generative Adversarial Networks, Social Bias, Gender Equality\\
		%\textbf{Topics:} AI, risks, society
		%Association for the Advancement of Artificial Intelligence\\
		%2275 East Bayshore Road, Suite 160\\
		%Palo Alto, California 94303\\
	}
	
	\maketitle
	
	\begin{abstract}
		The use of synthetic data generated by Generative Adversarial Networks (GANs) has become quite a popular method to do data augmentation for many applications. While practitioners celebrate this as an economical way to get more synthetic data that can be used to train downstream classifiers, it is not clear that they recognize the inherent pitfalls of this technique. In this paper, we aim to exhort practitioners against deriving any false sense of security against data biases based on data augmentation. To drive this point home, we show that starting with a dataset consisting of head-shots of engineering researchers, GAN-based augmentation ``imagines'' synthetic engineers, most of whom have masculine features and white skin color (inferred from a human subject study conducted on Amazon Mechanical Turk). This demonstrates how  biases inherent in the training data are reinforced, and sometimes even amplified, by GAN-based data augmentation; it should serve as a cautionary tale for the lay practitioners.

	\end{abstract}
	
	\begin{figure*}[!t]
		\centering
		\includegraphics[width=\textwidth]{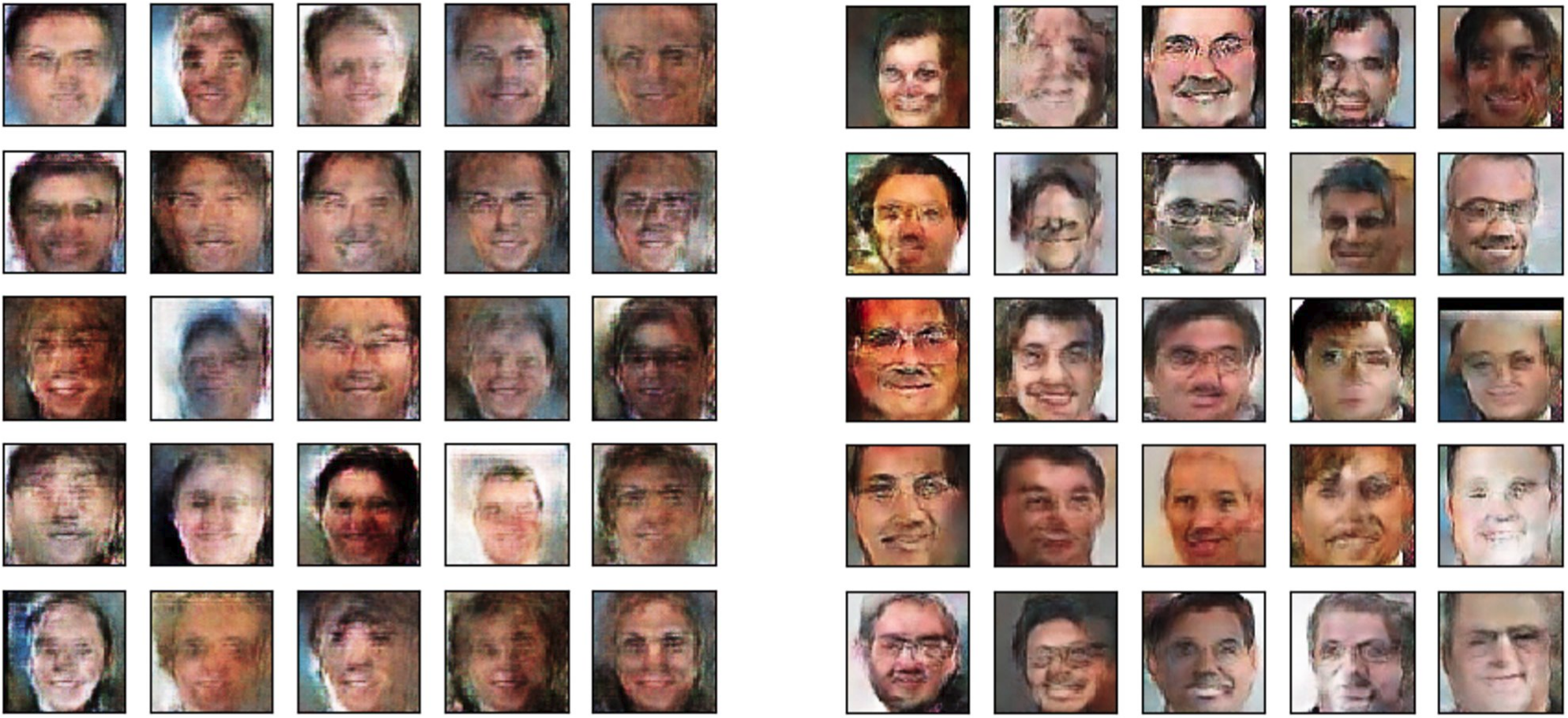}
		%\vspace{-2em}
		\caption{Synthetic faces generated by the DCGAN when asked to imagine faces of researchers in engineering (resemblance to anybody living or dead is purely coincidental). To the left is a sample of images after $20$ epochs, and to the right is after $925$ epochs. As training time increases, the images generated by the GAN improve in quality. When human subjects were asked to label each image as having features related to gender or skin color, we noticed the presence of both gender and racial bias. Furthermore, we noticed that although the GAN was expected to mimic the training data distribution $p_{data}$, it eventually not only propagated, but also amplified the bias; in order to fool the discriminator into classifying a generated image as that of a researcher in engineering, the generator learned to make these faces have more masculine features with lighter skin tones.}
		\label{fig:scenario1}
	\end{figure*}
	
	\section{Introduction}
	\begin{flushright}
		{\em Induction extends our expectation,\\
			not our experience.}
	\end{flushright}
	
	\noindent 
	Deep Learning techniques have emerged as the state-of-the-art technology for many machine learning problems, especially in the area of Computer Vision and Natural Language Processing. Their success on publicly available datasets for particular tasks has made it a lucrative solution for practitioners in multiple areas.
	Due to the large number of parameters that need to be learned for most deep learning architectures, the techniques necessitate training on massive amounts of data. Legal constraints, privacy, and scarcity of data can make this task too expensive to achieve. Data augmentation methods have sought to alleviate this problem. Recently, the approach of using Generative Adversarial Networks (GAN) to generate synthetic training data \cite{antoniou2017data} has gained tremendous popularity with discussions on the advantages and disadvantages of this type of data~\cite{forbes-synthetic}. For computer vision, these techniques are often preferred over traditional data augmentation techniques like image transformation (such as rotation or translation of the images in the the original dataset). Although GANs generate images that seem like samples from the same distribution as the training data, one cannot trivially identify any set of images in the training data from which a particular example was generated. This engenders the belief among practitioners that GAN-based techniques generate comprehensive  new data and thus, augmentation with these, would lead to more robust classifiers.
	
	With that hope, these techniques have been used across fields to train classifiers to identify medical issues \cite{baur2018melanogans,frid2018synthetic,wu2018conditional}, detect chip failures or anomalies \cite{lim2018doping}, recognize emotions \cite{wassersteineeg}, generating patient records \cite{choi2017generating}, novel design of drugs~\cite{jing2018deep}, etc. Researchers have also highlighted the advantages of using such augmentation techniques in regards to producing neural network classifiers that are less partial to specific network architectures and quantities of training data, and so require less fine-tuning of hyper-parameters by hand \cite{hernandez2018further}.
	
	Despite the increasingly widespread use of GAN-based data augmentation, the pitfalls in using this technique are not widely recognized; this work aims to show that GAN-based data augmentation can perpetuate biases inherent in the original data. Since GANs, at best, can only capture the distribution of the training data, GAN-generated data can be no more immune to data biases than the real-world data it attempts to mimic. Toward this investigation, we show that starting with a dataset consisting of head-shots of researchers in engineering, GAN-based augmentation ``imagines'' synthetic engineers, most of whom have masculine features and white skin color (as evaluated via human studies). This reveals GAN-based data augmentation can propagate the biases inherent in the training data, and sometimes, even amplify them. While this caution may not be particularly revelatory to serious machine learning researchers, it is worth emphasizing considering the widespread use of these techniques by non-expert practitioners.
	
	In the rest of the paper, we (1) describe the details of how the original training data was gathered and pre-processed, (2) explain how we generate the synthetic data using a DCGAN~\cite{radford2015unsupervised}, and (3) measure how biases pertaining to gender and skin color perpetuate (and even worsen) from the training data to the synthetically generated data. To ensure impartiality, these measures were derived from human-subject study on Amazon Mechanical Turk platform (\url{https://www.mturk.com/}).  
	
	%For many supervised learning tasks, the training dataset that is built using historical data, might have some biases towards certain features the learner ought not learn. Unfortunately, it proves to be a challenge to specify to inscrutable systems a set of features not to influence classification decisions. For example, a classifier that looks at r\'esum\'es to call candidates for an interview ought to ignore gender, even if it proves to be a distinguishing feature in historically labeled data. In light of recent news \cite{amazon-ai-recruiting-tool}, this is often easier said than done.
	
	%We warn users of such techniques that augmenting real-world data with GAN-generated synthetic data, which can be no less biased, will do no more than give false security in those biases by providing more evidence of them.
	
	%shifted the \iffalse one to come after end{document}

	\section{Approach}
	
	In this section, we first describe the distribution that a generator has to learn. We will use this as a representative example throughout the paper for illustration of our ideas. This is followed by a description of the technicalities of the GAN we use and the process of gathering data for training and testing.
	
	%One of the objectives of this paper is to see if these important characteristics learned about $p_{data}$ are features based on which we don't want new synthetic data to be generated (like gender, race, skin color etc.) because it would imply that classifiers learned on this data also learn the biases.

	\subsection{Model and Task Description}
\begin{figure*}[!t]
		\centering
		\includegraphics[width=\textwidth]{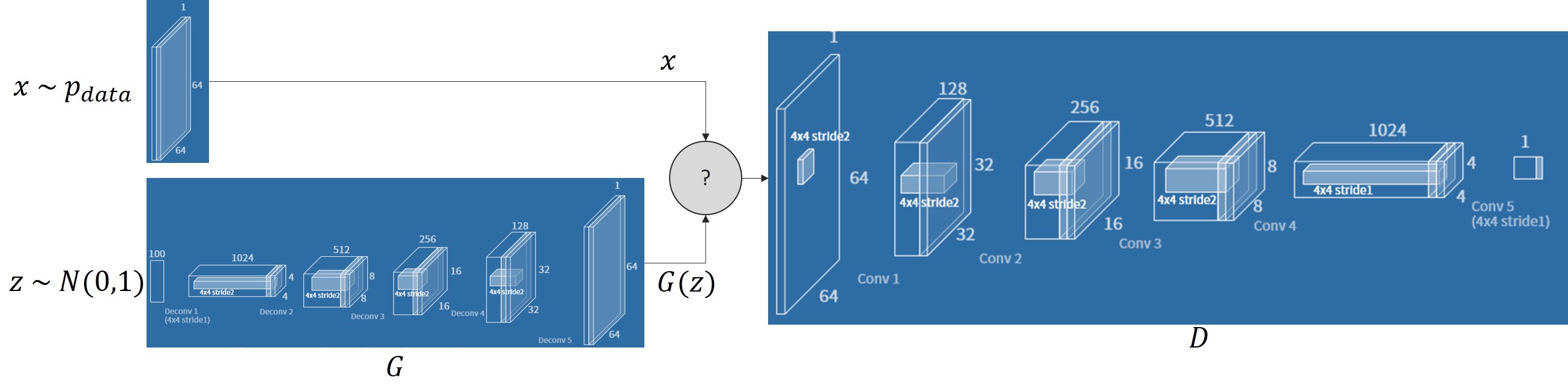}
		%\vspace{-2em}
		\caption{Deep Convolution Generative Adversarial Network (DCGAN) used to generate the faces of engineering researchers.}
		\label{fig:dcgan}
	\end{figure*}

	The main task of this work is to generate the image of a human face that looks like a researcher in engineering. We use the Deep Convolutional GAN~\cite{radford2015unsupervised} architecture shown in Fig. \ref{fig:dcgan} where the discriminator and the generator are denoted by $D$ and $G$ respectively. For all practical purposes, the discriminator $D$ learns an approximation of the true distribution given the finite set of training samples -- this approximation is denoted as $p_{data}$ -- from which images corresponding to faces of engineering researchers are generated.\\
	
	The generator $G$ generates a $64\times 64$-pixel image, denoted as $G(z)$, starting from randomly generated noise, represented as $100$-element vector $z$ sampled from the normal distribution $\mathcal{N}(0,1)$, with the hope $D$ will not be able to distinguish between an image sampled from $p_{data}$ and the generated image $G(z)$.
	For GANs, this is done by optimizing a loss function that $G$ aims to minimize and $D$ aims to maximize giving it the flavor of a min-max game:
	
	\[
	\min_{G} \max_{D} \mathbb{E}_{x \sim p_{data}} [\log D(x)] + \mathbb{E}_{z \sim \mathcal{N}(0,1)} [\log (1-D(G(z)))]
	\]%
	
	\noindent The first term is proportional to the accuracy of $D$ in classifying the actual data as real. The second term is proportional to the accuracy of $D$ in classifying the synthetically generated images $G(z)$ as fake. Therefore, the discriminator aims to maximize the overall term, while the generator aims only to minimize the second term (note that the first term is independent of $G$). Over time, the generator learns the most important characteristics of $p_{data}$ well enough to make $D$ believe the generated images are sampled from $p_{data}$.
	
	\subsection{Data Collection and Processing}
	%Towards this goal of investigating whether a generator learns bias on prohibitive features, 
	To investigate the propagation of biases in the synthetically generated data, we crawled two different data sources and created a single dataset which consists of head-shots of researchers in engineering.
	First, we scraped images of engineering researchers from eight universities in U.S.A.%nited States.
	%that were top ranked for their graduate programs in engineering as per the U.S. News \& World Report.
	\footnote{We refrain from naming the specific universities in this writeup because we believe that the data is indicative of the presence of gender and racial biases among hired faculty in  universities in general as opposed to only in the  ones we crawled.} %We do however plan to make our complete data set available freely to other researchers for replication.}
	%such as \emph{Georgia Tech}, \emph{Penn State}, \emph{University of Maryland}, \emph{UT Austin}, etc. 
	%These universities were selected from the list of top ranked universities in engineering graduate programs according to U.S. News \& World Report.
	
	All these universities had publicly available faculty directories which included head-shots of engineering researchers. The crawler scraped a subset of the image elements from a selected webpage, disregarding non-researcher images such as university logos and default stock images. The omission of non-professor images had human verification. This process helped us gather a total of $4,211$ images. We then gathered a set of images of researchers in Artificial Intelligence (AI) by following a snowball approach to crawl the profile pictures of these researchers from their Google Scholar profile pages (\url{https://scholar.google.com/}). This process helped us gather a dataset of $15,632$ images, thus making the size of the combined dataset $19,843$.
	
	Discrepancies in the data presented a myriad of challenges that we had to examine to have proper input data for the Deep Convolution Generative Adversarial Network.
	All images were not studio head-shots and some of the pictures seemed to be clicked in conditions ranging form natural environments to low-light conditions. This introduced three potential concerns -- (1) the presence of certain elements in the background or the surroundings could lead a generator $G$ generate features relating to those surroundings as opposed to the person's face in order to convince the discriminator that the generated image comes from $p_{data}$ (and these issues would be hard to debug given deep learning models are inscrutable) (2) the images had differences in regards to intensity of lighting (eg. studio {\em vs.} natural lighting), and (3) we needed images of uniform dimensions to give as input to the discriminator, which in our case has to be $64\times 64$ to match the dimensions of $G(z)$ (as mentioned earlier). 
	
	To ensure (1) didn't happen, we cropped the exact part of the photo containing the face along with a very small part of the background. For this task, we used a feature-based \textit{Histogram of Oriented Gradients} (HOG) face detector \cite{Dalal05} which first converted the image to black and white. It then broke the image into squares of $16\times 16$ pixels, to calculate all pixel gradients within those squares (which are the changes in the intensity of each pixel from those of its neighbors). It then averaged the results of all of them, obtaining the most predominant gradient for each square. Note that this approach also made the face detection resilient to changes in brightness, ensuring that problem (2), mentioned above, was also addressed. Finally, the detector compares the image with a known HOG face pattern and finds where it is, returning its coordinates. In cases when the module could not identify a face (which mostly happened for the default google scholar images), we discarded the image. In the case a face was recognized, we noticed that the cropped face was oftentimes too tightly cropped, missing multiple features such as ears, neck or frontal hair, which we felt were essential for properly recognizing gender. Thus, we decided to get a slightly broader region of their coordinates. Lastly, the final steps left were only to crop the selected face region and resize it to $64\times 64$ pixels, solving issue (3).
	
	\begin{figure*}[!t]
		\centering
		\begin{minipage}{.48\textwidth}
			\centering
			\includegraphics[width=\columnwidth]{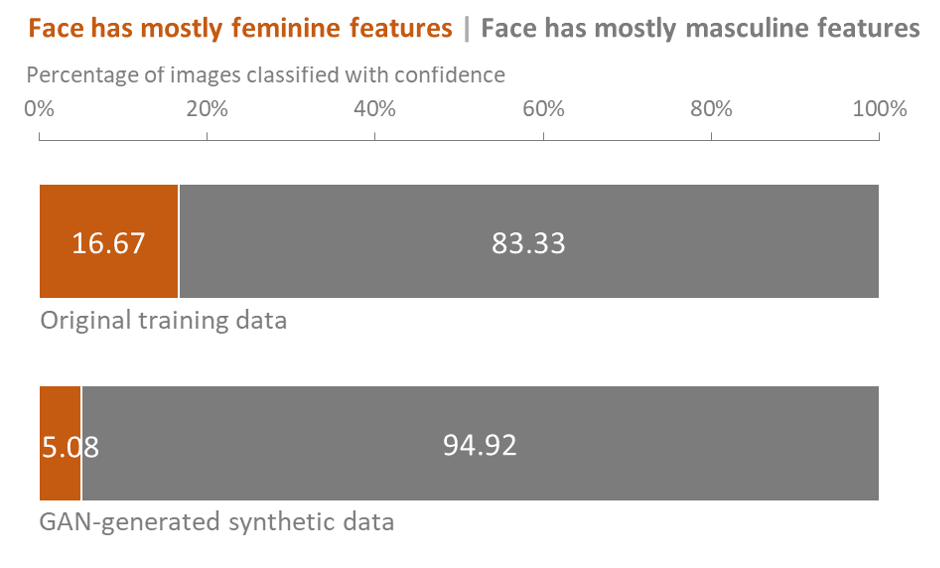}
			%\vspace{-2em}
			\caption{The percentage of faces classified as having feminine features, by at least eight human subjects, decreased from $16.67\%$ in the original dataset to $5.08\%$ in the synthetically generated dataset after 925 epochs.}
			\label{fig:mf}
		\end{minipage}
		~~~~
		\begin{minipage}{.48\textwidth}
			\centering
			\includegraphics[width=0.99\columnwidth]{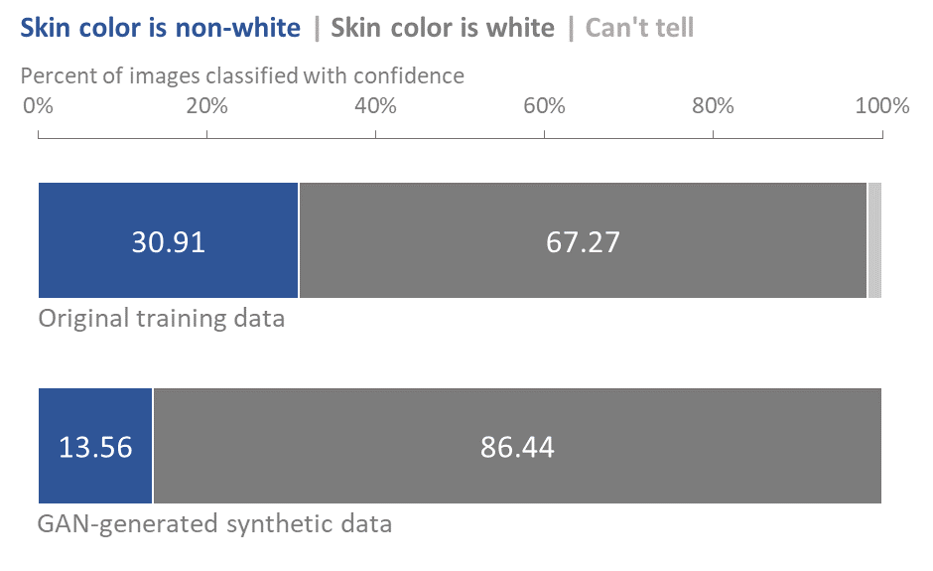}
			\caption{The percentage of faces classified as appearing non-white, by at least eight human subjects, decreased from $30.91\%$ in the original dataset to $13.56\%$ in the synthetically generated dataset after 925 epochs.}
			\label{fig:nwn}
		\end{minipage}
	\end{figure*}
	
	Once the aforementioned preprocessing steps were applied to our datasets, $6209$ images were discarded; our final dataset is comprised of $13634$ images. Almost all of the images rejected were of researchers on Google Scholar who chose their profile pictures with no human faces; we noticed that the main reason behind this dropped set of images was a great amount of the data we scraped from Google Scholar used the default profile picture. The face detector simply could not recognize any faces from this avatar.
	
	\section{Experimental Results}
	
	In this section, we describe the studies with human subjects that help us analyze the generated images. The results show that the `imagined' faces of researchers in engineering not only perpetuate the racial and gender bias we saw in the original data, but exacerbate it.
	%and in fact eventually propagates the biases over prohibitive features of the training data further.
	
	\subsection{Analysis of Generated Images}
	
	% \begin{figure}[!t]
	% \centering
	% \includegraphics[width=0.8\columnwidth]{mf.png}
	% %\vspace{-2em}
	% \caption{The percentage of faces which had feminine features, as per the majority of human subjects, went down from $16.67\%$ in the original dataset to $5.08\%$ in the synthetically generated dataset.}
	% \label{fig:mf}
	% \end{figure}
	
	We analyzed potential gender and racial biases that the GAN may have learned by conducting studies using the human subjects. We chose a set of $120$ images-- of which $60$ were randomly sampled from the original data\footnote{We do not show images randomly sampled from the training set as they are of real engineering researchers, who might not be comfortable disclosing their identity, especially in the context of our paper.} (denoted as $x$) and the other $60$ were randomly sampled from images generated by the GAN (denoted as $G(z)$) after 925 epochs.
	%Of these, we sample $90$ images to get the later set of images `A: I believe this 30-image sampling should be random'.
	We then conducted four human study tasks as follows:
	
	\begin{itemize}
		\item[\texttt{T1a}] Human subjects were asked to select the most appropriate option for an image $x$ sampled from $p_{data}$ with the following options: a) face mostly has masculine features, b) face mostly has feminine features, and c) neither of the above are true.
		\item[\texttt{T1b}] Human subjects were asked to perform a task identical to {\tt T1a}, but for a synthetically generated image $G(z)$.
		\item[\texttt{T2a}] Human subjects were asked to select the most appropriate option for an image $x$ sampled from the training data $p_{data}$ from the list of following options: skin color is non-white, skin color is white, and can't tell.
		\item[\texttt{T2b}] Human subjects were asked to perform a task identical to {\tt T2a} but for a synthetically generated image $G(z)$.
	\end{itemize}
	
	\begin{figure*}[!t]
		%\centering
		%\begin{minipage}{.48\textwidth}
		\centering
		\includegraphics[width=0.9\textwidth]{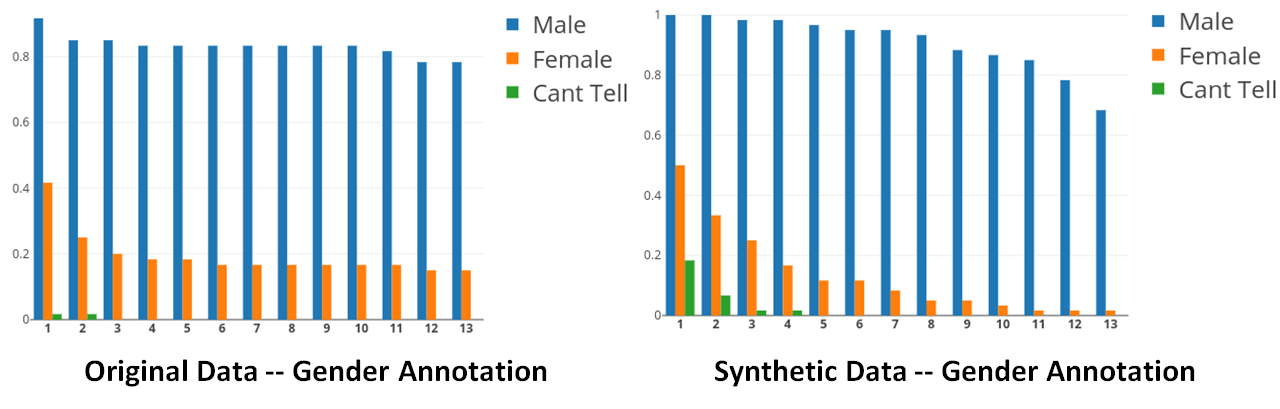}
		%\vspace{-2em}
		\caption{The number of images labeled as masculine, feminine, or neither, changes as the threshold number of votes required to categorize an image into a particular category increase from $1$ to $15$.}
		\label{fig:turkerDisagGender}
		%\end{minipage}
	\end{figure*}
	\begin{figure*}[!t]
		%\begin{minipage}{.48\textwidth}
		\centering
		\includegraphics[width=0.9\textwidth]{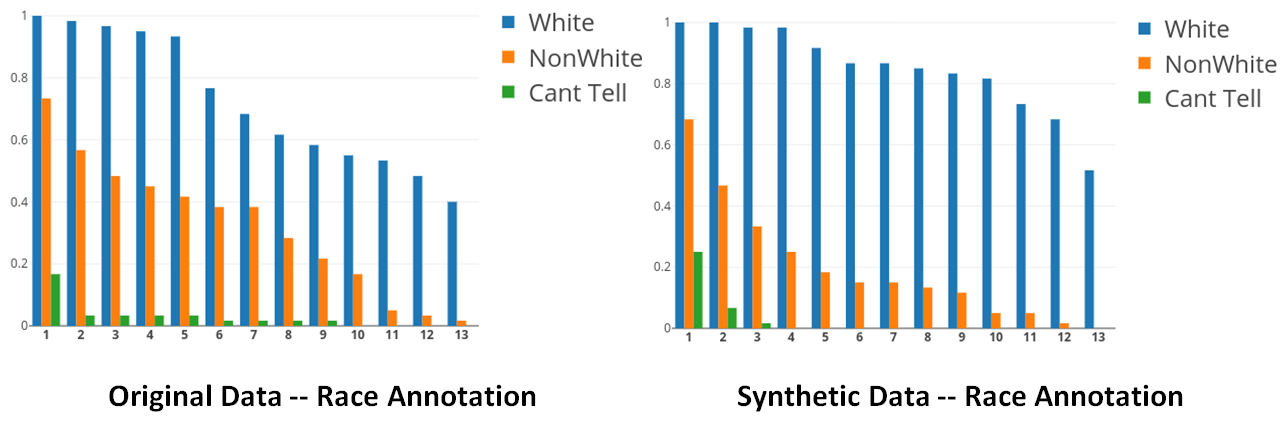}
		%\vspace{-2em}
		\caption{The number of images labeled as white, non-white, or can't tell, changes as the threshold number of votes required to categorize an image into a particular category increase from $1$ to $15$.}
		\label{fig:turkerDisagRace}
		%\end{minipage}
	\end{figure*}
	
	To ensure that the data was of high quality, we paid \$1 to each of the $60$ workers on MTurk where, each task or a HIT (Human Intelligence Tasks) takes approximately four minutes. Each MTurk worker had a master qualification, which indicates that they have high reputation earned by completing multiple tasks that have been approved by the task requesters previously. Each worker was given a set of $32$ images-- of which $30$ were from either the original dataset ($x$) or completely generated, i.e. $G(z)$ (but no mix-and-match), and two were images of stationary objects -- a bottle and a chair -- for which the answers to all the tasks were obvious (`neither of the above' for {\tt T1a} and {\tt T1b} and `can't tell' for {\tt T2a} and {\tt T2b}). These last two images helped prune meaningless spam obtained from any human subjects who finished a HIT without actually paying attention to the question, or used a bot that gives a same answer to all the questions in the HIT. In our experiment, this method helped us to identify and prune one such human subject's given data. 
	
	%Given that GANs are expected to learn the (approximate) distribution of $p_{data}$ given a finite number of samples, we hypothesize the following:
	
	% \begin{tcolorbox}
	% \begin{itemize}
	% \item[{\tt H1}] The percentage of faces that have feminine (or masculine) features will remain the same for {\tt T1a} and {\tt T1b}.
	% \item[{\tt H2}] The percentage of faces that have non-white (or white) skin color will remain the same for {\tt T2a} and {\tt T2b}.
	% \end{itemize}
	% \end{tcolorbox}%
	
	Our aim is to assess how the bias pertaining to the features associated with gender and race in the initial dataset changes or remains approximately the same in the synthetic dataset. For the purposes of our analysis, we followed a majority-voting metric to categorize an image as having a feature if at least eight of the $15$ human subjects labeled it as such. We plot the results for {\tt T1a} and {\tt T1b} in Figure \ref{fig:mf}. Quite interestingly, we found the percentage of images which had mostly masculine features increased from $83.33\%$ in the original data to $94.92\%$ in the synthetically generated data, while the percentage of images which had mostly feminine features decreased from $16.67\%$ to $5.08\%$. These metrics show an increase in the gender bias when a generator is asked to imagine an engineering researcher. 
	%We discuss the ramifications of this after shedding light on the second hypothesis.
	
	We also noticed (as shown in Figure \ref{fig:nwn}), again using the concept of majority voting, for tasks {\tt T2a} and {\tt T2b}, the number of images that had a face with a white skin tone increased from $67.27\%$ in the original dataset to $86.44\%$ in the synthetically generated dataset. {\em This result 
		%also invalidates our hypothesis {\tt H2}, and in fact 
		showcases that the generator learns to bias the generated faces as those of people with lighter skin tones when asked to imagine an engineering researcher}. From the results, we noticed that the situation is worse because the synthetically generated data not only propagated, but increased the bias toward the minority population in our original data, i.e. faces with feminine features and darker skin tones.
	
	\iffalse
	\begin{figure}[!t]
		\centering
		\includegraphics[width=\columnwidth]{CombinedGenderPC.png}
		%\vspace{-2em}
		\caption{The bar plot represents}
		\label{fig:turkerDisagGender}
	\end{figure}
	
	\begin{figure}[!t]
		\centering
		\includegraphics[width=\columnwidth]{CombinedRacePC.png}
		%\vspace{-2em}
		\caption{The bar plot represents}
		\label{fig:turkerDisagRace}
	\end{figure}
	\fi
	
	%The hypothesis that we had set out to evaluate were to show practitioners the use of GANs (which are expected to capture the distribution $p_{data}$) could lead to perpetuating the biases pertaining to high-level features such as gender and race. 

	%The invalidation of both hypotheses leads to the conclusion that even though a GAN is expected to capture the input distribution $p_{data}$, which it probably is doing even in our case with respect to the low-level input features (such as pixels of an image), it is unable to ensure that it captures the distribution with respect to high-level features such as gender and race. Furthermore, we see, from our experiments, that it reduces the representation of these high-level features in the generated data for the minority populations (humans with feminine facial features and darker skin tones).
	
	We plot the confidence metrics based on the number of votes for each option by the human subjects in Figures \ref{fig:turkerDisagGender} and \ref{fig:turkerDisagRace}. The y-axis indicates the number of images for which at least $n$ number of people (plotted on the x-axis) categorized as a particular option (each of which is indicated by a color). It is interesting to notice that in Figure~\ref{fig:turkerDisagGender}, for every image in both the original and the synthetically generated data there was at least one person ($n = 1$) who thought that the face `had mostly masculine features'. As the number of people who voted for a certain class increases along the x-axis, the number of images for which the crowd approached a unanimous decision ($n = 13$) also decreases. We further observe that the reduction in the confidence (measured by the number of votes for an option) is higher for `having mostly feminine features' as opposed to `having mostly masculine features' -- there was {\em no image} in which $13$ or more people though that the face mostly had mostly female features. 
	
	In Figure \ref{fig:turkerDisagRace}, we notice a similar trend where the confidence of the crowd (as measured by the number of votes for a particular option) reduces at a faster rate for the option stating that the skin color of the person is non-white, quickly diminishing to under $20\%$ beyond $n = 10$, while the number of votes for the option `skin color is white' does not reduce drastically for the synthetic data. This also seems to be a harder task than the previous one as the number of images with high number of human subjects agreeing on an option decreases as we move from left to right. The threshold value $n$ for which we categorize an image as having a feature could have been any value. \textit{This highlights that even if we did not consider majority voting, we would have seen the same results showing that 
		%the data bias two hypothesis {\tt H1} and {\tt H2} were invalidated, thereby indicating that 
		GANs are not merely perpetuating our biases towards gender, skin color etc. but rather exacerbating them}. 
	
	We measured the inter-annotator agreement to estimate how well the human subjects agreed with one another on a specific task in our scenario using Fleiss Kappa~\cite{FleissKappa} and Krippendorff's alpha~\cite{RichAlpha}. Inter-annotator agreement (denoted as $\kappa$~\cite{FleissKappa} here) is a measure of how well two (or more) annotators can make the same annotation decision for a certain category. When calculating this metric, we noticed that when annotating gender specific features, the value of this metric was very high ($\kappa$ is 0.765) when compared to annotating the skin color of a given image ($\kappa$ is 0.326). This implies that human subjects agreed more with one another when annotating images for presence of gender specific features but agreed less when annotating an image with respect to the skin color. Considering that we observed this for both the original and the synthetically generated data, we feel that the task of choosing the skin color proved to be more difficult for the human subjects than labeling gender-specific features.
	The $\kappa$ values also highlighted that the human subjects agreed more with each other for the original data than the synthetic data. We feel this may be due the quality of images generated by GANs that are not at par with the ones in the original dataset. 
	
	%This, as we now discuss, is something that can have major ramifications when GAN generated data is used for data augmentation. 
	
	\section{Discussion and Conclusion}
	
	%Thus, using GANs in for augmenting data in classification tasks where there are extraneous features on which we want to prohibit classification can be disastrous, especially if using an inscrutable classifier. For example, if a bank decided to give loan to a person on not based on their images using a deep learning classifier and were using data augmentation using GANs, it would still be having bias towards people with gender or skin color, which is be prohibited by the Federal Trade Commission \cite{credit-no-discriminate}. Hence, even if GANs are proving to be the state-of-the-art for data augmentation, we caution users to be aware that they will surely propagate (and may even worsen, as in our case) the social biases.
	
	Beyond implications about social issues, this work also highlights the implications of the study when GAN-based data augmentation is used for tasks like generating images of chips with defects or lab samples of malignant conditions relating to a particular disease. There seems to be a false sense of security that the GANs will generate {\em new} data that picks the expected semantic features (in this case, parts of the image that actually represent the `defect' of a chip or `abnormality' in a lab image) more. Also, if the kind of defects in a newly manufactured chip or effects of a new virus on the medical image changes, the classifier trained on augmented data will be worse off at detecting these because of similar reasons highlighted earlier in the section.
	
	%\section{Conclusions}
	
	In this paper, we caution practitioners to be more aware of the issue that the use of GANs for data-augmentation 
	%as a panacea for solving the problem of data availability 
	might inadvertently perpetuate biases present in the data. We show when a state-of-the-art GAN is asked to imagine faces of a researcher in engineering, it generates faces that seem to have a strong bias for masculine facial features and white skin color. Although we expected perpetuation of biases, via studies conducted on Amazon's MTurk, we found the GAN-generated data even exacerbated the bias present in the original data in regard to gender and skin color. Thus, the use of such data for augmentation makes it difficult for future data to correct the biases (because there is more biased data to offset); the bias of yesterday on gender or race influences the decisions of today and tomorrow. Lastly, we note that while we focused on perpetuation of obvious and troubling social biases, the application should be extrapolated to other fields; GAN-based data-augmentation can perpetuate any type of extraneous bias. This should give pause to the widespread use of GAN-based data augmentation in medical and anomaly-detection domains. 
	
	% It must be noted that there may have been features not discussed above which could have created a noticeable disparity in images between the two datasets. For example, the images of engineering researchers were curated from head shots on university directory and Google Scholar profiles; this may entail better studio lighting and 
	
	%\clearpage
	\balance
	\bibliographystyle{aaai}
	\bibliography{ref}
	
\end{document}